\documentclass[sigconf]{acmart}
\AtBeginDocument{%
  }

\copyrightyear{2026}
\acmYear{2026}
\setcopyright{cc}
\setcctype{by}
\acmConference[KDD '26]{Proceedings of the 32nd ACM SIGKDD Conference on Knowledge Discovery and Data Mining V.1}{August 09--13, 2026}{Jeju Island, Republic of Korea}
\acmBooktitle{Proceedings of the 32nd ACM SIGKDD Conference on Knowledge Discovery and Data Mining V.1 (KDD '26), August 09--13, 2026, Jeju Island, Republic of Korea}
\acmPrice{}
\acmDOI{10.1145/3770854.3785698}

\acmISBN{979-8-4007-2258-5/2026/08}
\usepackage{multirow}
\usepackage{enumitem}
\usepackage{balance}

\setlist[itemize]{leftmargin=12pt}
\begin{document}

\title{SynSym: A Synthetic Data Generation Framework \\ for Psychiatric Symptom Identification}

\author{Migyeong Kang}
\affiliation{%
  \institution{Sungkyunkwan University}
  \country{Seoul, Republic of Korea}
}
\email{gy77@g.skku.edu}

\author{Jihyun Kim}
\affiliation{%
  \institution{Sungkyunkwan University}
  \country{Seoul, Republic of Korea}
}
\email{wlgus01110@g.skku.edu}

\author{Hyolim Jeon}
\affiliation{%
  \institution{Sungkyunkwan University}
  \country{Seoul, Republic of Korea}
}
\email{gyfla1512@g.skku.edu}

\author{Sunwoo Hwang}
\affiliation{%
  \institution{Sungkyunkwan University}
  \country{Seoul, Republic of Korea}
}
\email{sunwoo1357@g.skku.edu}

\author{Jihyun An}
\affiliation{%
  \institution{Samsung Medical Center}
  \country{Seoul, Republic of Korea}
}
\email{jh85.an@samsung.com}

\author{Yonghoon Kim}
\affiliation{%
  \institution{Omnicns}
  \country{Seoul, Republic of Korea}
}
\email{yhkim525@omnicns.com}

\author{Haewoon Kwak}
\affiliation{%
  \institution{Indiana University}
  \country{Bloomington, IN, United States}
}
\email{haewoon@acm.org}

\author{Jisun An}
\affiliation{%
  \institution{Indiana University}
  \country{Bloomington, IN, United States}
}
\email{jisun.an@acm.org}

\author{Jinyoung Han}
\authornote{Corresponding author.}
\affiliation{%
  \institution{Sungkyunkwan University}
  \country{Seoul, Republic of Korea}
}
\email{jinyounghan@skku.edu}

\renewcommand{\shortauthors}{Migyeong Kang et al.}

\begin{abstract}
    Psychiatric symptom identification on social media aims to infer fine-grained mental health symptoms from user-generated posts, allowing a detailed understanding of users' mental states. However, the construction of large-scale symptom-level datasets remains challenging due to the resource-intensive nature of expert labeling and the lack of standardized annotation guidelines, which in turn limits the generalizability of models to identify diverse symptom expressions from user-generated text. To address these issues, we propose \textsc{SynSym}, a synthetic data generation framework for constructing generalizable datasets for symptom identification. Leveraging large language models (LLMs), \textsc{SynSym} constructs high-quality training samples by (1) expanding each symptom into sub-concepts to enhance the diversity of generated expressions, (2) producing synthetic expressions that reflect psychiatric symptoms in diverse linguistic styles, and (3) composing realistic multi-symptom expressions, informed by clinical co-occurrence patterns. We validate \textsc{SynSym} on three benchmark datasets covering different styles of depressive symptom expression. Experimental results demonstrate that models trained solely on the synthetic data generated by \textsc{SynSym} perform comparably to those trained on real data, and benefit further from additional fine-tuning with real data. These findings underscore the potential of synthetic data as an alternative resource to real-world annotations in psychiatric symptom modeling, and \textsc{SynSym} serves as a practical framework for generating clinically relevant and realistic symptom expressions.

\end{abstract}



\maketitle

\section{Introduction}
Mental health disorders represent a global public health concern, with nearly half of the world's population expected to experience at least one mental disorder over the course of their lifetime~\cite{mcgrath2023age}. With the increasing use of social networks, people frequently share their psychological experiences online, offering new opportunities for language-based mental health assessment~\cite{de2013predicting}. Although early natural language processing (NLP) research has primarily focused on predicting the risk of mental disorders~\cite{de2016discovering, shen2017detecting, tadesse2019detection, lee2022detecting}, there is increasing interest in identifying fine-grained psychiatric symptoms reflected in the language of users~\cite{nguyen2022improving, zhang2022symptom, cai2023depression, kang2024cure}. Symptom identification aims to recognize individual psychiatric symptoms, such as anhedonia, insomnia, or depressed mood, based on clinical instruments such as the PHQ-9 (Patient Health Questionnaire)~\cite{kocalevent2013standardization} and DSM-5 (Diagnostic and Statistical Manual of Mental Disorders)~\cite{american2013diagnostic}. Beyond disorder-level classification, this task enables early detection of high-risk individuals and supports personalized intervention~\cite{zhang2023phq}.

Despite its clinical importance, psychiatric symptom identification remains a challenging task primarily due to the inherent difficulties in constructing high-quality training datasets---particularly the complexity of the annotation process for symptom labels~\cite{mowery2015towards}. One key difficulty is that multiple psychiatric symptoms frequently co-occur within a single user-generated post, necessitating fine-grained multi-label annotation guidelines~\cite{zhang2023phq}. Another issue is the lack of standardized labeling protocols. While disorder-level classification can leverage standardized instruments such as the DSM-5 or PHQ-9, symptom-level annotation depends on subjective judgment, leading to inconsistencies and low inter-annotator agreement~\cite{milintsevich2024your}. These challenges indicate that the construction of reliable training datasets for symptom identification requires well-defined annotation protocols and the close involvement of mental health professionals throughout the annotation process.

These structural challenges have been widely revealed in most existing studies on psychiatric symptom identification. For example, the number of samples per symptom class is extremely small in most prior work; some rare symptoms are represented by as few as 100 instances, making it difficult for models to learn the diversity of symptom expressions~\cite{gupta2022learning, yadav2020identifying, zhang2022symptom, yadav2023towards}. Moreover, the reliability of the annotation remains a key obstacle. Many publicly available datasets have been labeled by non-experts~\cite{gupta2022learning, yadav2020identifying, zhang2022symptom}, and a recent study has raised concerns about the consistency and clinical validity of these annotations. For instance, \citet{milintsevich2024your} re-annotated a subset of the PRIMATE dataset~\cite{gupta2022learning}, originally labeled by crowd workers, with the help of a mental health professional, and found that the agreement with the original labels was remarkably low ($\kappa = 0.09$). In this way, existing datasets on psychiatric symptom identification are mostly small in scale and suffer from limited label reliability, making it difficult for models to learn diverse and clinically meaningful symptom expressions. In addition, because most datasets are constructed from a single platform or linguistic style~\cite{gupta2022learning, yadav2020identifying, zhang2022symptom}, models often struggle to generalize to new platforms or unseen styles of symptom expressions.

To address these limitations, we introduce \textsc{SynSym}, an LLM-based synthetic data generation framework for symptom identification. Given only a list of symptom classes and their brief descriptions, \textsc{SynSym} automatically produces diverse linguistic expressions that reflect the presence of each symptom, with minimal manual effort. The framework consists of four key stages. First, it expands each symptom into fine-grained sub-concepts to comprehensively reflect the diversity of symptom expressions. Second, it generates synthetic symptom expressions in two distinct styles, clinical and colloquial, to capture the stylistic variability observed across platforms and user populations. Third, it generates synthetic examples that reflect multiple co-occurring symptoms, guided by clinical knowledge of symptom co-occurrence patterns. Lastly, it performs quality control by using an LLM to validate whether the generated expressions appropriately align with the intended symptom labels. To the best of our knowledge, \textsc{SynSym} is the first framework to leverage large language models for generating synthetic data for psychiatric symptom identification. By addressing the challenges of limited training data and unreliable labels, \textsc{SynSym} facilitates the construction of high-quality and diverse corpora that support more robust and generalizable symptom identification across diverse settings. 

To evaluate the effectiveness of \textsc{SynSym}, we conduct experiments on three widely used benchmark datasets for symptom identification. Given that prior research on symptom identification has primarily focused on depressive disorders, we concentrate on three datasets that cover depressive symptoms. Experimental results demonstrate that models trained solely on the synthetic data generated by \textsc{SynSym} perform comparably to those trained on real data, and benefit further from additional fine-tuning with real data. Further experiments on other psychiatric conditions indicate that our framework generalizes beyond depressive disorders, showcasing its extensibility. These findings highlight \textsc{SynSym}'s ability in generating high-quality data that captures real-world symptom expressions and generalizes across diverse linguistic styles. All resources including the synthetic dataset and code are publicly available to support future research\footnote{\url{https://github.com/gyeong707/SynSym-KDD-2026}}.

\section{Related Work}

\subsection{Psychiatric Symptom Identification on Social Media}
Social networks have emerged as valuable data sources for mental health research, as they allow users to express their psychological states anonymously, enabling early detection of mental conditions~\cite{yates2017depression, chen2018mood, gaur2019knowledge, lee2020cross, lee2024detecting}. Recent research has focused on symptom-level prediction based on clinical diagnostic tools, such as the PHQ-9~\cite{kocalevent2013standardization} and DSM-5~\cite{american2013diagnostic}. These studies tried to detect fine-grained psychiatric symptoms from social media text~\cite{farruque2021explainable, yao2021extracting, yadav2023towards, farruque2024depression}, usually structuring the task to resemble the process of completing clinical questionnaires~\cite{li2024zero, gupta2022learning, yadav2020identifying}. The predicted symptoms are often integrated into disorder classification models to improve interpretability and diagnostic precision~\cite{zhang2022symptom, chen2023detection, kang2024cure}. Symptom-level representations can support personalized intervention strategies, such as identifying high-risk individuals and informing treatment planning~\cite{yadav2020identifying}, which highlights the importance of building reliable and accurate systems for symptom identification in mental health applications.

\subsection{Datasets for Psychiatric Symptom Identification}
Several datasets have been proposed to support psychiatric symptom identification. D2S is a Twitter-based dataset for detecting depressive symptoms~\cite{yadav2020identifying}, labeled according to the PHQ-9 criteria. Similarly, PRIMATE~\cite{gupta2022learning} is a Reddit-based dataset designed to identify depressive symptoms aligned with the PHQ-9 items in users' posts. \citet{zhang2022symptom} constructed a dataset called PsySym, aiming at detecting multiple mental disorders along with their corresponding symptoms. The dataset was collected from Reddit and annotated using a scheme based on the DSM-5 diagnostic criteria. Despite these efforts in building datasets for identifying psychiatric symptoms, most existing resources are relatively small in scale and rely on crowd-sourced annotations from user-generated posts~\cite{gupta2022learning, yadav2020identifying, zhang2022symptom}, raising concerns about the reliability of their labels~\cite{milintsevich2024your}. In addition, existing datasets are often collected from a single platform and reflect narrow expression patterns, hindering the development of generalizable models across various settings. For instance, in D2S~\cite{yadav2020identifying}, users tend to express their symptoms in an implicit and figurative manner, while in PsySym~\cite{zhang2022symptom}, symptoms are described more directly using clinical terminology. These limitations emphasize the need for training datasets that are not only extensive and reliable but also sufficiently diverse to represent symptom manifestations across different social media platforms and expression styles.

\begin{figure*}[h]
    \centering
    \includegraphics[width=\textwidth]{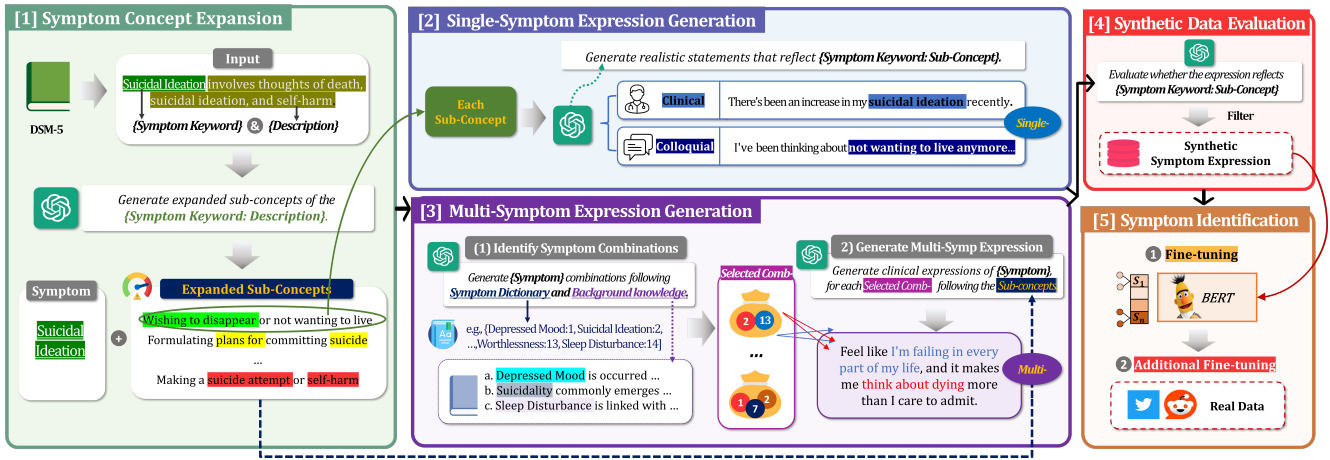}
    \caption{The \textsc{SynSym} framework consists of four stages: 
        (1) \textit{Symptom Concept Expansion}, which decomposes each symptom into fine-grained sub-concepts; 
        (2) \textit{Single-Symptom Expression Generation}, which produces expressions of individual symptoms in two linguistic styles: clinical and colloquial; 
        (3) \textit{Multi-Symptom Expression Generation}, which simulates realistic co-occurring symptom expressions based on clinical knowledge; and 
        (4) \textit{Synthetic Data Evaluation}, which assesses whether the generated expressions accurately reflect real clinical symptoms. 
        The resulting synthetic dataset is then used to train multi-label classification models for identifying psychiatric symptoms from user-generated text.}
    \label{fig:framework}
\end{figure*}

\subsection{LLM-Based Synthetic Data Generation for Mental Disorder Detection}
Mental health data often contains sensitive personal information, such as personal relationships and detailed medical histories, posing limitations on collecting large-scale training corpora~\cite {ive2022leveraging}. As a promising alternative, recent research has explored synthetic data generation using large language models in mental disorder detection. For instance, \citet{vedanta2024psychsynth} used GPT-4o and Nemotron to generate realistic narratives of fictional individuals with anxiety and depression. \citet{ghanadian2024socially} proposed a method for generating synthetic suicidal ideation data grounded in psychological literature. Similarly, \citet{kang2024synthetic} and \citet{aygun2023use} synthesized interview-style dialogues to augment the DAIC-WOZ dataset, a widely used depression corpus. These studies showed that synthetic data can improve training diversity, mitigate class imbalance, and enhance model performance when combined with real data. However, they predominantly focused on binary disorder-level classification and generated synthetic data tailored to a single platform~\cite{vedanta2024psychsynth, ghanadian2024socially, kang2024synthetic, aygun2023use}, limiting the evaluation of generalizability across diverse expression styles. In contrast, we propose a synthetic data generation framework for multi-label psychiatric symptom identification, which can ensure generalizability across diverse platforms and expression styles.

\section{\textsc{SynSym}:  A Framework for Psychiatric Symptom Expression Generation}
We present \textsc{SynSym}, a synthetic data generation framework for generalizable psychiatric symptom identification. \textsc{SynSym} generates synthetic symptom expressions using an off-the-shelf LLM without requiring any task-specific fine-tuning. While \textsc{SynSym} is designed to be applicable across a wide range of psychiatric conditions, in this study, we focus on depressive symptoms as a representative case. An overview of the \textsc{SynSym} framework is illustrated in Figure~\ref{fig:framework}, and all prompt templates used in the framework are described in Appendix~\ref{appendix:prompts_framework}. In the following subsections, we describe an overview of each stage in the framework, the motivation behind each design choice, and implementation details, including prompting strategies.

\subsection{Symptom Concept Expansion}
    \noindent\textbf{Overview.} The first stage of synthetic data generation in \textsc{SynSym} is the expansion of psychiatric symptom concepts into more specific and fine-grained sub-concepts. Given a high-level symptom keyword (e.g., \textit{Suicidal Ideation}) and a brief clinical description, this step outputs a set of expanded sub-concepts that represent diverse, concrete ways the symptom can manifest in real-world scenarios. In this study, we extract symptom categories from the DSM-5 diagnostic criteria~\cite{american2013diagnostic} for major depressive disorder.
    \vspace{0.3em}
    
    \noindent\textbf{Motivation.} This step is designed to enhance the diversity of synthetic symptom expressions generated by LLMs. Manually crafting detailed descriptions for each psychiatric symptom can be both time-consuming and expertise-dependent. While brief descriptions may provide some guidance to LLMs, it is not easy to extract the full range of symptom expressions using the short descriptions. To address this, we leverage the LLM's domain-specific knowledge to expand each high-level symptom label into fine-grained sub-concepts automatically. This allows us to establish more diverse and concrete entry points for generation, enabling the model to generate varied and realistic expressions in the next stage, while minimizing manual effort.
    \vspace{0.3em}

    \noindent\textbf{Implementation.} We designed a structured prompt that provides the LLM with a high-level symptom keyword and its brief clinical description. The model is then instructed to output at least ten fine-grained sub-concepts that reflect diverse real-world manifestations of the symptom. As a result, the model produces a comprehensive set of sub-concepts for each symptom keyword, capturing diverse manifestations and varying degrees of clinical severity. These outputs are then manually reviewed and refined by the authors to eliminate redundancy and ensure clear separation between related symptoms.

    \begin{table}[]
    \caption{Summary of three benchmark datasets and our synthetic dataset. Compared to real-world datasets, \textsc{SynSym} provides substantially more training samples per symptom class, alleviating the issue of class imbalance commonly observed in psychiatric symptom corpora.}
    \renewcommand{\arraystretch}{1.2}
    \centering
    \resizebox{1.0\linewidth}{!}{
        \begin{tabular}{lcccc}
        \hline
        \multicolumn{1}{c}{\textbf{Statistic}} & \textbf{PsySym} & \textbf{PRIMATE} & \textbf{D2S} & \textbf{\textsc{SynSym}} \\ \hline
        \textbf{\# Samples} & 1,433 & 2,003 & 1,717 & 18,254 \\
        \textbf{\# Symptom Classes} & 14 & 9 & 9 & 14 \\ \hline
        \textbf{Avg. Post Length} & 18.5 & 299.3 & 17.9 & 23.2 \\
        \textbf{Avg. Symptoms per Sample} & 1.4 & 3.5 & 1.2 & 1.6 \\ \hline
        \textbf{Avg. Samples per Class} & 143.5 & 779.3 & 227.4 & 2119.5 \\
        \textbf{Max. Samples per Class} & 348 & 1,679 & 527 & 4,829 \\
        \textbf{Min. Samples per Class} & 51 & 195 & 46 & 1,186 \\ \hline
        \end{tabular}}
    
        \label{table:dataset_statistics}
    \end{table}
    
\subsection{Single-Symptom Expression Generation}
    \noindent\textbf{Overview.} The second stage of \textsc{SynSym} generates realistic symptom expressions by leveraging the previously constructed sub-concepts. Given a symptom keyword and its corresponding sub-concept description, this step produces two stylistic variants to reflect the diversity of symptom expressions observed in a real-world context as follows.
    \vspace{0.3em}

    \begin{itemize}
        \item \textit{Clinical Style}: Formal and objective statements that incorporate clinical terminology, resembling phrasing used in medical or diagnostic documentation.
        \item \textit{Colloquial Style}: Informal and first-person descriptions that reflect how individuals express symptoms on social media platforms in a direct and explicit manner.
    \end{itemize}

    \noindent\textbf{Motivation.} This dual-style design stems from a practical limitation observed in LLMs. When prompted to generate direct statements, LLMs often avoid clinically explicit or sensitive terms, such as \textit{loss of libido}, \textit{self-harm}, or \textit{suicidal thoughts}, and instead produce vague or softened alternatives (e.g.,\textit{ ``I wish to disappear''}). Such avoidance limits the model's exposure to critical symptom indicators, weakening its ability to detect high-risk or clinically relevant expressions in real-world user content. To address this, we enforce the generation of clinical expressions to ensure sufficient coverage of essential diagnostic language in the synthetic data. 
    \vspace{0.3em}

    \noindent\textbf{Implementation.} Each prompt generated 10 symptom expressions in a first-person perspective, based on a symptom keyword and its sub-concept. To prevent the LLM from avoiding direct or clinically explicit language, we explicitly instructed the model to produce 5 statements in a clinical style and 5 in a colloquial style. This process was repeated five times for each sub-concept, resulting in a total of 50 synthetic expressions per sub-concept. Representative examples of the two styles are shown in Appendix~\ref{appendix:synthetic_dataset}.

\subsection{Multi-Symptom Expression Generation}
    \noindent\textbf{Overview.} This stage focuses on simulating realistic expressions that reflect the co-occurrence of multiple psychiatric symptoms within a single user post. It involves two steps: (1) identifying symptom co-occurrence combinations, and (2) generating multi-symptom expressions that naturally describe these combinations. Firstly, we generate symptom co-occurrence pairs from the full list of target symptoms using the LLM (e.g., Suicidal Ideation and Worthlessness). In this stage, to ensure that the generated combinations reflect realistic clinical scenarios, we provide the LLM with background knowledge that describes typical comorbidity patterns in depression. This knowledge is constructed by reviewing prior clinical literature on depressive symptom co-occurrence~\cite{quinn2023relations, gijzen2021suicide, mullarkey2019using, fried2016good}. Subsequently, based on these symptom combinations and their corresponding sub-concepts, we prompt the LLM to generate multi-symptom expressions that naturally reflect the joint manifestation of each pair.
    \vspace{0.3em}

    \noindent\textbf{Motivation.} Symptom identification is typically implemented as a multi-label classification task, as user-generated posts often contain multiple co-occurring symptoms. In this setting, models are trained to learn not only the individual expressions of each symptom but also the underlying co-occurrence patterns between the target classes. Consequently, training with samples that reflect unrealistic or arbitrarily mixed symptom combinations can lead to unnatural expressions and spurious correlations, which may hinder model generalization and reliability. To mitigate this issue, we design multi-symptom expressions based on clinically grounded co-occurrence patterns, ensuring that the synthetic data reflects realistic symptom dependencies observed in real-world cases.
    \vspace{0.3em}

    \noindent\textbf{Implementation.} To construct co-occurring symptom combinations, the LLM was provided with (1) a full list of target symptom keywords and (2) background knowledge describing symptom co-occurrence patterns, extracted from prior clinical literature. Based on this input, the LLM was prompted to generate plausible combinations, each consisting of 2 to 5 symptoms with varying severity levels. This process was repeated until a target number of diverse combinations was reached (e.g., 10,000 combinations). In the subsequent stage, the model takes as input (1) a symptom combination and (2) the full list of sub-concepts for the involved symptoms, and generates expressions that describe multiple symptoms within a single post. Similar to the previous stage, we generate both clinical and colloquial expressions for each combination.

\subsection{Synthetic Expression Evaluation}
    After generating both single- and multi-symptom expressions, we conducted an evaluation step to ensure the reliability of symptom labels. Specifically, we prompted the model to assess how well each synthetic expression reflects its corresponding target symptoms. Each expression was rated on a 5-point scale based on its semantic alignment with the intended symptoms. Expressions receiving a score of 2 or lower were removed. This step enables quality control of the synthetic dataset by ensuring that only expressions aligned with their intended symptoms are retained.


    \begin{table*}[]
    \caption{Representative examples of the \textit{Suicidal Ideas} symptom from the three benchmark datasets. \textsc{PsySym} contains concise, clinically grounded expressions, whereas \textsc{D2S} uses more figurative and indirect language. \textsc{PRIMATE} includes significantly longer posts that often encapsulate multiple symptom cues within a single entry (e.g., both \textit{Suicidal Ideas} and \textit{Feeling Down} are labeled in the presented example).}
    \renewcommand{\arraystretch}{1.0}
    \centering
    \small
    \resizebox{1.0\linewidth}{!}{
    \begin{tabular}{r|l}
    \hline
    \multicolumn{1}{c|}{\textbf{Dataset}} & \multicolumn{1}{c}{\textbf{Example of Expression}} \\ \hline
    \textbf{PsySym \cite{zhang2022symptom}} & I have begun to have suicidal ideation again, something that I have not experienced in a very long time. \\ \hline
    \textbf{PRIMATE \cite{gupta2022learning}} & \begin{tabular}[c]{@{}l@{}}So Much Pain In The World I read Tara Condell's suicide note and it really hit home for me. I have been struggling \\ with suicidal thoughts for years. Sometimes it seems like the little things in life aren't enough to make up \\ for all the suffering I see in the world. I can't understand how someone can have 4 houses, take 12 cruises a year \\ and fly around in private jets when everyone else is dying. How can we be this selfish as a species? \\ Darwinism no longer controls us, it's survival of the wealthy, which I know I will never be.\end{tabular} \\ \hline
    \textbf{D2S \cite{yadav2020identifying}} & My mind is quite literally killing me. \\ \hline
    \end{tabular}}

    \label{tab:benchmark_examples}
    \end{table*}

    \begin{table}[]
        \caption{Results of expert validation on the two core components of \textsc{SynSym}. Each item was rated for clinical validity by two licensed psychiatrists on a 5-point Likert scale. Inter-rater agreement represents the proportion of items rated identically by both experts.}
        \renewcommand{\arraystretch}{1.1}
        \centering
        \resizebox{1.0\linewidth}{!}{
        \begin{tabular}{c|c|c|c}
        \hline
        \textbf{Component} & \textbf{\begin{tabular}[c]{@{}c@{}}Expert 1\\ \end{tabular}} & \textbf{\begin{tabular}[c]{@{}c@{}}Expert 2\end{tabular}} & \textbf{\begin{tabular}[c]{@{}c@{}}Agreement\end{tabular}} \\ \hline
        \textbf{Expanded Sub-Concepts} & 4.61 & 4.57 & 94.86\% \\ \hline
        \textbf{Synthetic Expressions} & 4.99 & 5.00 & 99.66\% \\ \hline
        \end{tabular}}
        \label{table:expert_validation}
    \end{table}

\section{Experimental Setting}
In this section, we describe the experimental setup, including the synthetic dataset constructed using \textsc{SynSym}, the benchmark datasets used for evaluation, and the training and evaluation settings.

\subsection{Synthetic Dataset}
    \subsubsection{\textbf{Statistical Descriptions}}
    We constructed a synthetic dataset for depressive symptom identification using the proposed \textsc{SynSym} framework. We defined 14 target symptom categories and their brief clinical descriptions based on the PsySym dataset~\cite{zhang2022symptom}, which provides DSM-5--based symptom annotations for multiple mental disorders, including depression. All stages of the framework were implemented using GPT-4o. On average, 18 sub-concepts were generated per symptom, resulting in 18,254 synthetic samples, comprising 12,621 single-symptom and 5,633 multi-symptom entries. Table~\ref{table:dataset_statistics} presents a statistical overview of the generated dataset.

    \subsubsection{\textbf{Expert Validation}}
    We conducted an additional expert validation to assess the clinical validity of the synthetic data generated by \textsc{SynSym}. Two certified psychiatrists participated in this evaluation process. We focused on two core components of \textsc{SynSym} that require clinical domain expertise: 
    \vspace{0.1em}

    \begin{itemize}
        \item \textit{Expanded Sub-Concepts}: A total of 253 fine-grained sub-concepts derived from 14 depressive symptoms. Experts were asked to evaluate whether each sub-concept appropriately represented the corresponding high-level symptom, without exaggeration or distortion.
    \vspace{0.2em}
        \item \textit{Synthetic Symptom Expressions}: A set of 300 expressions selected via stratified sampling to ensure balanced coverage across diverse symptom labels. Experts rated whether each expression accurately conveyed the meaning of its corresponding symptom(s) and their associated sub-concepts.
    \end{itemize}
    
    \noindent Table~\ref{table:expert_validation} reports the average scores from both experts. The evaluation results indicate strong clinical validity for both components. For sub-concepts, the average rating exceeded 4.6, suggesting that the fine-grained symptom expressions generated by \textsc{SynSym} are clinically relevant to depressive disorders. A small number of low-scoring items were identified as being more indicative of other psychiatric conditions—for example, "inattention" is more characteristic of ADHD, although it can still be observed in depressive episodes. For synthetic expressions, the average ratings approached the maximum score of 5.0, demonstrating that the generated symptom expressions accurately reflect the intended symptoms and their sub-concepts.

   \begin{table*}[h!]
        \caption{Overall performance on three benchmark datasets for psychiatric symptom identification. We adopted MentalBERT~\cite{ji2021mentalbert} as the backbone encoder for models trained using \textsc{SynSym}-generated data. 
        \textsc{SynSym}\textsubscript{w/o Real} refers to models trained solely on synthetic data, while \textsc{SynSym}\textsubscript{with Real} refers to models initialized with \textsc{SynSym}\textsubscript{w/o Real} and further fine-tuned on real data. The highest score for each column is shown in bold. For PRIMATE, due to the longer average post length compared to the synthetic dataset, individual sentences were evaluated separately and their predicted symptoms were aggregated for \textsc{SynSym}\textsubscript{w/o Real}.} 
        \renewcommand{\arraystretch}{1.2}
        \centering
        \footnotesize
        \resizebox{1.0\linewidth}{!}{
        \begin{tabular}{rcccccc}
        \hline
        \multicolumn{1}{c}{\multirow{2}{*}{\textbf{Model}}} & \multicolumn{2}{c}{\textbf{PsySym \cite{zhang2022symptom}}} & \multicolumn{2}{c}{\textbf{PRIMATE \cite{gupta2022learning}}} & \multicolumn{2}{c}{\textbf{D2S \cite{yadav2020identifying}}} \\ \cline{2-7} 
        \multicolumn{1}{c}{} & Macro-Rec ↑ & Macro-F1 ↑ & Macro-Rec ↑ & Macro-F1 ↑ & Macro-Rec ↑ & Macro-F1 ↑ \\ \hline
        \textbf{BERT} \cite{devlin2018bert} & 0.738 ± 0.005 & 0.814 ± 0.004 & 0.625 ± 0.024 & 0.646 ± 0.021 & 0.555 ± 0.046 & 0.600 ± 0.046 \\
        \textbf{DeBERTa} \cite{he2020deberta} & 0.747 ± 0.007 & 0.796 ± 0.003 & 0.604 ± 0.054 & 0.614 ± 0.049 & 0.535 ± 0.041 & 0.563 ± 0.045 \\
        \textbf{MentalBERT} \cite{ji2021mentalbert} & 0.730 ± 0.005 & 0.811 ± 0.004 & 0.629 ± 0.038 & 0.643 ± 0.031 & 0.554 ± 0.033 & 0.603 ± 0.030 \\ \hline
        \textbf{GPT4o-ZSL} \cite{achiam2023gpt} & 0.774 ± 0.004 & 0.808 ±0.005 & 0.569 ± 0.001 & 0.567 ± 0.000 & 0.631 ± 0.001 & 0.588 ± 0.003 \\
        \textbf{GPT4o-COT} \cite{wei2022chain} & 0.775 ± 0.004 & 0.807 ± 0.002 & 0.567 ± 0.000 & 0.568 ± 0.000 & \textbf{0.642 ± 0.003} & 0.592 ± 0.003 \\
        \textbf{GPT4o-PS} \cite{wang2023plan} & 0.755 ± 0.002 & 0.808 ± 0.005 & 0.529 ± 0.002 & 0.543 ± 0.001 & 0.578 ± 0.003 & 0.579 ± 0.004 \\ \hline
        \textbf{\textsc{SynSym}}\textsubscript{w/o Real} & 0.732 ± 0.001 & 0.778 ± 0.003 & \textbf{0.712 ± 0.017} & 0.557 ± 0.011 & 0.525 ± 0.023 & 0.518 ± 0.022 \\
        \textbf{\textsc{SynSym}}\textsubscript{with Real} & \textbf{0.798 ± 0.012} & \textbf{0.830 ± 0.006} & 0.645 ± 0.022 & \textbf{0.650 ± 0.016} & 0.588 ± 0.025 & \textbf{0.614 ± 0.023} \\ \hline
        \end{tabular}}
        \label{table:overall_performance}
    \end{table*}

\subsection{Benchmark Datasets}
    \subsubsection{\textbf{Characteristics}}
        We evaluate the effectiveness of \textsc{SynSym} with three benchmark datasets for depressive symptom identification. A statistical overview of each dataset is presented in Table~\ref{table:dataset_statistics}. For D2S, due to issues such as the deletion of original tweets, only 1,717 posts were collected, fewer than the originally reported 3,738. All datasets exhibit noticeable class imbalance, and some symptom classes contain fewer than 100 instances. The three datasets differ significantly in their linguistic expression styles, as illustrated in the representative examples shown in Table~\ref{tab:benchmark_examples}.

        \begin{itemize}
            \item \textbf{PsySym}~\cite{zhang2022symptom}: A Reddit-based dataset labeled with DSM-5 symptoms, where posts typically describe psychiatric symptoms in a direct and explicit manner.
            
            \item \textbf{PRIMATE}~\cite{gupta2022learning}: A Reddit dataset annotated with PHQ-9 symptoms, featuring long, narrative-driven posts that often detail personal struggles.
            
            \item \textbf{D2S}~\cite{yadav2020identifying}: A Twitter-based dataset labeled with PHQ-9 symptoms. Posts are short and frequently employ figurative or sarcastic language.
        \end{itemize}

    \subsubsection{\textbf{Remapping Labels}}
        The benchmark datasets used in our experiments follow different annotation schemes, which necessitated aligning the label sets in certain experiments (e.g., cross-dataset evaluation in Section~\ref{sec:generalizability}). While PHQ-9 includes 9 depressive symptoms, DSM-5 covers 14 categories that provide a more comprehensive scope of depressive symptomatology. Therefore, we performed a remapping process to align DSM-5-based labels with the PHQ-9 annotation scheme where appropriate. Detailed mapping rules and procedures are provided in Appendix~\ref{appendix:remapping_labels}.

    \subsubsection{\textbf{Rewriting for D2S}}
        In one experimental setting, we applied an additional rewriting step to the D2S dataset to address a stylistic mismatch with the \textsc{SynSym}-generated data. While datasets like PsySym and PRIMATE typically contain direct and explicit symptom descriptions (e.g., “I’m so tired I can’t get out of bed”), D2S is characterized by more figurative or abstract expressions (e.g., “My thoughts feel like a sinking ship”), which were excluded from our synthetic data generation due to concerns that such expressions may blur class boundaries and increase the risk of false positives. Therefore, when evaluating the D2S dataset on models trained solely on synthetic data, we used an LLM to rewrite the D2S samples into clearer and more clinically oriented expressions. This rewriting step was applied only in this specific setting, while the original D2S data was used in all other experimental conditions.

\subsection{Model Training}
We evaluate the effectiveness of \textsc{SynSym} under different training settings for psychiatric symptom identification.

    \subsubsection{\textbf{Training with Synthetic Data}}  We trained a multi-label classifier for psychiatric symptom identification using the generated synthetic dataset. Considering its demonstrated effectiveness in symptom identification tasks~\cite{zhang2022symptom, milintsevich2024your}, we adopted MentalBERT~\cite{ji2021mentalbert} as the backbone encoder. We fine-tuned MentalBERT on the synthetic data generated by \textsc{SynSym} to learn representative symptom-related expression patterns. The constructed dataset was split 8:2 for training and validation, respectively.
    
    \subsubsection{\textbf{Additional Training with Real Data}}
    Although training on synthetic data enabled the model to capture generalizable symptom expressions, it is not sufficient to align with real-world datasets annotated by humans, due to inconsistencies in annotation schemes and underlying diagnostic frameworks. Consequently, models trained solely on synthetic data may fail to match the labeling conventions of human-annotated datasets. To address this gap, we further fine-tuned the model trained on synthetic data using real-world datasets collected from social media platforms (e.g., Reddit and Twitter). This additional step helps the model adapt to dataset-specific labeling practices and better reflect the nuances of human annotations.

\subsection{Baselines}
We designed two types of baselines to reflect realistic deployment scenarios for psychiatric symptom identification. All models were evaluated under a multi-label classification setting, where the objective is to predict the presence or absence of multiple symptoms from a single input text. 
    
    \subsubsection{\textbf{BERT-based Approaches}}
    BERT-based models have been widely adopted as strong baselines in prior research on psychiatric symptom identification~\cite{gupta2022learning, zhang2022symptom}. In our experiments, we considered three representative BERT-based models: \textbf{BERT}~\cite{devlin2018bert}, a general-purpose language model pretrained on large-scale English corpora; \textbf{DeBERTa}~\cite{he2020deberta}, which introduces disentangled attention mechanisms and has demonstrated strong performance across diverse NLP benchmarks; and \textbf{MentalBERT}~\cite{ji2021mentalbert}, a domain-adapted variant further pretrained on mental health–related social media data to better capture psychiatric expressions. All models were fine-tuned on individual benchmark datasets (e.g., D2S and PRIMATE) using a standard supervised training pipeline and evaluated on the corresponding held-out test sets.

    
    \subsubsection{\textbf{LLM-based Approaches}}
    In scenarios where labeled training data is limited or unavailable, large language models can be directly applied via prompting. We evaluated three GPT4o-based approaches under a zero-shot setting, without any task-specific fine-tuning. \textbf{GPT4o-ZSL} performs direct multi-label classification by prompting the model with a task description and symptom list~\cite{achiam2023gpt}. \textbf{GPT4o-COT} incorporates Chain-of-Thought prompting to encourage step-by-step reasoning before prediction~\cite{wei2022chain}. \textbf{GPT4o-PS} adopts a plan-and-solve strategy, in which the model first constructs an explicit reasoning plan and then derives the final predictions~\cite{wang2023plan}. We adopted a zero-shot setting for all LLM-based approaches to avoid the sensitivity of example selection in multi-label classification~\cite{hou2021few}.
    

    \subsection{Experimental Protocol}    
    \noindent\textbf{Data Partitioning.} To ensure fair comparison and reproducibility, we adopted standardized training and evaluation procedures across all experiments. For models trained on the synthetic dataset, we performed 5-fold cross-validation. Among the benchmark datasets, D2S and PRIMATE do not include predefined splits, so we also applied 5-fold cross-validation to these datasets. For PsySym, which provides official partitions, we followed its original split scheme. 
    \vspace{0.3em}

    \noindent\textbf{Randomness Control.} We trained each model using five different random seeds (\texttt{42} through \texttt{46}) to control for randomness and ensure robust performance estimation. The mean and standard deviation of performance were reported across all seeds and cross-validation folds. Further implementation details, including hyperparameter settings, are provided in Appendix~\ref{appendix:hyperparameters}.

\section{Experimental Results}    
    \subsection{Overall Performance}
        We first compare the overall performance of the baseline methods trained with \textsc{SynSym}-generated and the three benchmark datasets, as presented in Table~\ref{table:overall_performance}. Among the BERT variants, the three models showed only minor differences in performance, indicating that neither more advanced architectures nor domain-adapted pretraining consistently improves outcomes in this task. For the GPT4o zero-shot results, performance differences across prompting strategies were marginal between basic zero-shot (ZSL) and chain-of-thought (COT) prompting~\cite{wei2022chain}. However, the plan-and-solve (PS) strategy~\cite{wang2023plan}, which first constructs a reasoning plan before prediction, showed relatively lower performance, suggesting that multi-stage reasoning does not necessarily provide distinct benefits for symptom identification.

        Despite being trained solely on synthetic data, models trained with \textsc{SynSym}\textsubscript{w/o Real} achieved performance comparable to those trained on real-world benchmark datasets across all three datasets. This indicates that the synthetic data generated by \textsc{SynSym} are clinically meaningful and generalize well across diverse linguistic styles. When further fine-tuned on real data (\textsc{SynSym}\textsubscript{with Real}), the models demonstrated consistently improved performance, outperforming all baselines across most metrics, with the exception of recall on the D2S dataset. This lower performance gain can be attributed to the frequent use of figurative and indirect expressions in the D2S dataset. Such linguistic styles were deliberately excluded from our \textsc{SynSym} framework to avoid label ambiguity. While GPT4o-based models showed high recall on this dataset, leveraging their strong language understanding, this came at the expense of precision, often resulting in the over-diagnosis tendency noted in prior literature~\cite{sarma2025integrating}. In contrast, the \textsc{SynSym}\textsubscript{with Real} model demonstrated a superior balance between recall and precision, achieving higher F1-scores than most baselines. Detailed results of the performance analysis across individual symptom classes are provided in Appendix~\ref{appendix:experimental_result}.

        

    \begin{table}[t]
        \caption{The results of cross-dataset generalizability, where each row indicates the training dataset and each column indicates the test dataset. Gray-shaded cells indicate cases where the training and test datasets are the same.}
        \footnotesize
        \renewcommand{\arraystretch}{1.3}
        \centering
        \resizebox{1.0\linewidth}{!}{
        \begin{tabular}{l|cccccc}
        \hline
        \multicolumn{1}{c|}{\multirow{2}{*}{\textbf{Train}}} & \multicolumn{2}{c}{\textbf{Test: PsySym}} & \multicolumn{2}{c}{\textbf{Test: PRIMATE}} & \multicolumn{2}{c}{\textbf{Test: D2S}} \\ \cline{2-7} 
        \multicolumn{1}{c|}{} & Rec. ↑ & F1. ↑ & Rec. ↑ & F1. ↑ & Rec. ↑ & F1. ↑ \\ \hline
        \textbf{PsySym \cite{zhang2022symptom}} & \textcolor{gray}{0.730} & \textcolor{gray}{0.811} & 0.466 & 0.510 & 0.348 & 0.380 \\
        \textbf{PRIMATE \cite{gupta2022learning}} & 0.544 & 0.440 & \textcolor{gray}{0.629} & \textcolor{gray}{0.643} & 0.512 & 0.340 \\
        \textbf{D2S \cite{yadav2020identifying}} & 0.342 & 0.412 & 0.404 & 0.425 & \textcolor{gray}{0.554} & \textcolor{gray}{0.603} \\ \hline
        \textbf{Multi-Source} & 0.405 & 0.476 & 0.277 & 0.257 & 0.441 & 0.434 \\ \hline
        \textbf{\textsc{SynSym (ours)}} & \textbf{0.732} & \textbf{0.778} & \textbf{0.712} & \textbf{0.557} & \textbf{0.525} & \textbf{0.518} \\ \hline
        \end{tabular}}
        \label{table:generalizability}
    \end{table}

    \subsection{Generalizability across Datasets}
    \label{sec:generalizability}
    We evaluate the generalizability of models trained with different data sources, focusing on their ability to transfer across datasets with diverse linguistic characteristics. In addition to single-source training, where models are trained and tested on distinct individual datasets, we conducted a multi-source training experiment. In this setting, the model was trained on two of the three real-world datasets and evaluated on the remaining one. This approach aims to provide a more robust baseline by exposing the model to a broader range of linguistic and platform-specific expressions.

    Table~\ref{table:generalizability} summarizes the results across all combinations of train–test splits. Overall, models trained on individual real-world datasets exhibit limited generalization when applied to data from other datasets. This limitation is particularly evident between PsySym and D2S, which differ substantially in symptom expression styles. Models trained on multi-source data show relatively poor performance when evaluated on unseen datasets, suggesting that simple data aggregation is not an effective solution to the data scarcity in symptom identification. In contrast, models trained on \textsc{SynSym} data achieve consistently strong performance across all test sets, even without additional fine-tuning on real data, highlighting the strong cross-dataset generalizability of \textsc{SynSym}.

\subsection{Ablation Study}
    We conducted an ablation study to assess the contribution of each component in the \textsc{SynSym} framework, as shown in Table~\ref{table:ablation}. We progressively removed four core components: symptom expression evaluation \textit{(EV)}, clinical background knowledge of symptom co-occurrence patterns, which were derived from prior literature and used for generating multi-symptom expressions \textit{(CK)}, dual-style symptom expression generation\textit{(DU)}, and symptom concept expansion \textit{(SE)}. We observed that the impact of each component varied across datasets. For example, removing the dual-style generation led to a substantial performance drop on PsySym, while it slightly improved performance on D2S. Conversely, eliminating symptom concept expansion resulted in a sharp decline on D2S, but even slightly improved performance on PsySym. Nevertheless, the results indicate that each component of \textsc{SynSym} contributes to overall performance across datasets, with model performance generally degrading as more components are removed.

    \begin{table}[t]
        \caption{Results of the ablation study evaluating the contribution of each component in the SynSym framework. In the \textit{w/o CK} setting, symptom combinations were generated using only the LLM’s internal knowledge, without any clinically derived co-occurrence pattern.}
        \small
        \renewcommand{\arraystretch}{1.3}
        \centering
        \resizebox{1.0\linewidth}{!}{
        \begin{tabular}{l|llllll}
        \hline
        \multicolumn{1}{c|}{\multirow{2}{*}{\textbf{Model}}} & 
        \multicolumn{2}{c}{\textbf{PsySym}} & 
        \multicolumn{2}{c}{\textbf{PRIMATE}} & 
        \multicolumn{2}{c}{\textbf{D2S}} \\
        \cline{2-7}
        \multicolumn{1}{c|}{} & Rec. ↑ & F1. ↑ & Rec. ↑ & F1. ↑ & Rec. ↑ & F1. ↑ \\ \hline
        \textbf{\textsc{SynSym (ours)}} & \textbf{0.798} &\textbf{0.830} & \textbf{0.645} & \textbf{0.650} & \textbf{0.588} & \textbf{0.614} \\ \hline
        \small\hspace{0.1cm}\textit{w/o EV} & 0.785 & 0.821 & 0.633 & 0.640 & 0.583 & 0.610 \\
        \small\hspace{0.1cm}\textit{w/o EV + CK} & 0.770 & 0.817 & 0.630 & 0.633 & 0.572 & 0.612 \\
        \small\hspace{0.1cm}\textit{w/o EV + CK + DU} & 0.755 & 0.809 & 0.618 & 0.629 & 0.566 & 0.608 \\
        \small\hspace{0.1cm}\textit{w/o EV + CK + DU + SE} & 0.759 & 0.812 & 0.612 & 0.625 & 0.532 & 0.585 \\ \hline
        \end{tabular}}
        \label{table:ablation}
    \end{table}

    \subsection{Extensibility to Other Disorders}
\begin{figure}[]
        \centering
        \includegraphics[width=0.9\linewidth]{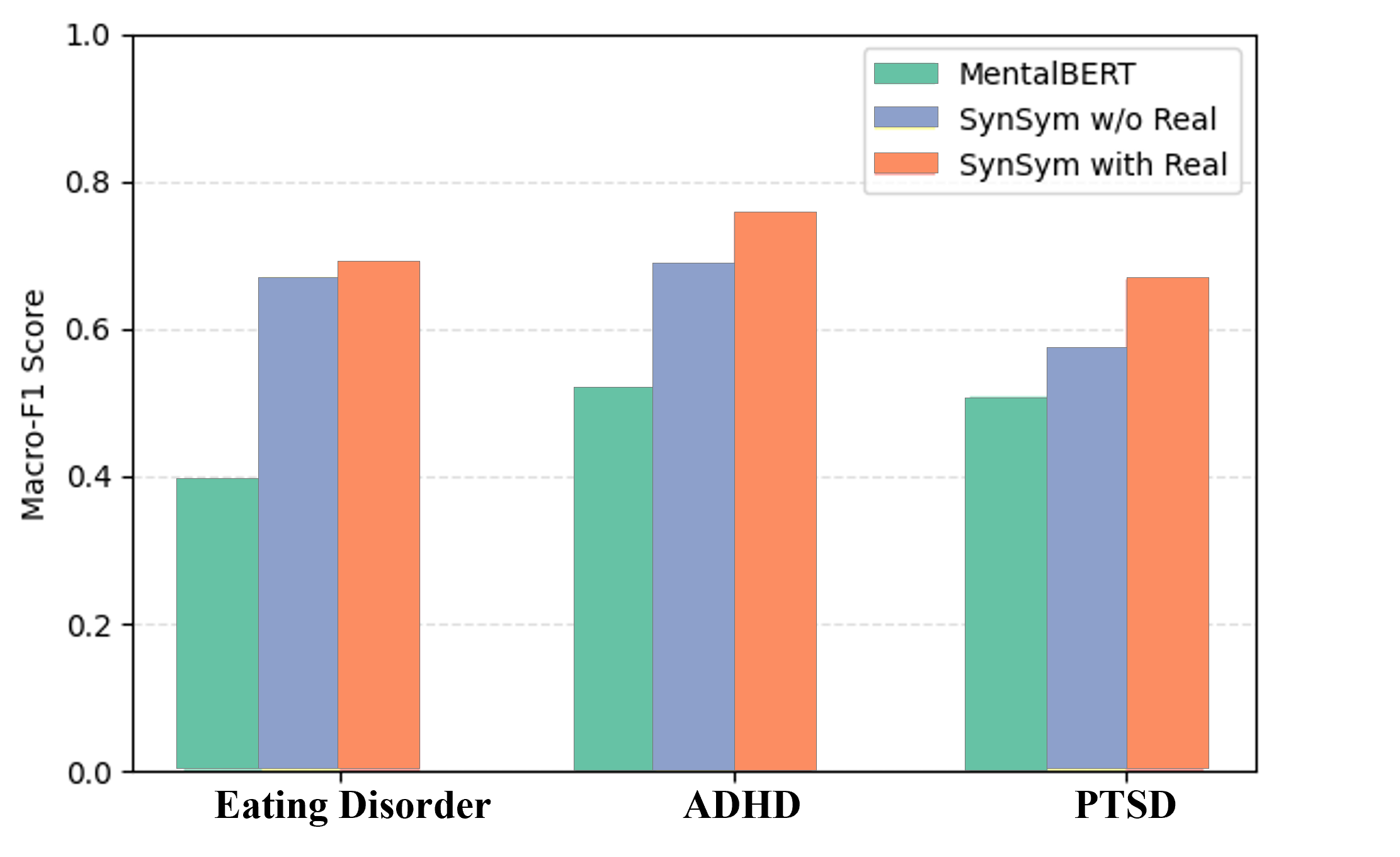}
        \caption{
        Extensibility experiment results across three mental health conditions. Bar heights represent the average Macro-F1 score across all their corresponding symptom classes.}
        \label{fig:scalability}
    \end{figure}

    We further evaluate its extensibility to other mental health conditions. The results are presented in Figure~\ref{fig:scalability}. Specifically, we conduct experiments on three additional disorders: post-traumatic stress disorder \textit{(PTSD)}, eating disorders \textit{(ED)}, and attention-deficit/hyperactivity disorder \textit{(ADHD)}, which are included in the PsySym dataset~\cite{zhang2022symptom}. In comparison, models trained solely on real data demonstrated limited ability to accurately identify symptom-related cues, due to the small number of training examples available for these disorders. In contrast, models trained on the \textsc{SynSym}-generated dataset show strong performance, even without access to real data. These results indicate that \textsc{SynSym} generalizes beyond depressive disorders to other psychiatric conditions and can serve as a foundational resource in scenarios where collecting training samples is difficult.

    \subsection{Comparison with Back-Translation}
    To examine whether the performance gains achieved by \textsc{SynSym} stem primarily from increasing the amount of training data, we compare our framework with back-translation, a widely adopted data augmentation technique that expands training data through paraphrasing. Specifically, we applied English--German--English back-translation to the training splits using Helsinki-NLP's OPUS-MT models. For each benchmark dataset, we first applied back-translation to the original training split and then augmented the training data by combining the back-translated samples with the original ones for model training. 
    
    The results show that incorporating back-translated data into the original training set led to modest performance improvements on PsySym, suggesting that subtle variations in expressions introduced by back-translation can contribute to performance gains. However, performance degraded on PRIMATE and D2S when back-translated data was added, compared to using real data alone. We attribute these declines to the amplification of labeling noise in PRIMATE and the distortion of figurative language in D2S through translation. In contrast, models trained with \textsc{SynSym} consistently achieved performance improvements across all three benchmarks. This suggests that obtaining clinically grounded and diverse symptom expressions is more critical for building generalizable symptom identification models than simply increasing dataset size through augmentation.

\section{Conclusion}
    This study introduces \textsc{SynSym}, a novel framework for generating synthetic data aimed at psychiatric symptom identification. Given a symptom keyword and its brief description as input, \textsc{SynSym} generates diverse and clinically relevant symptom expressions. Experiments conducted on three benchmark datasets for depression symptom identification demonstrate that models trained solely on \textsc{SynSym} data perform comparably to those trained on real-world data, and show further improvements when fine-tuned with a small amount of real data. These results suggest that \textsc{SynSym} can serve as an effective alternative or pretraining resource in real-world scenarios where high-quality labeled data is scarce.

    This work is significant in that it represents the first attempt to apply synthetic data to the task of symptom prediction and empirically demonstrates its effectiveness. However, several limitations remain. First, the experiments primarily focus on depressive disorders, necessitating future extension to other psychiatric conditions. Second, to preserve label reliability, \textsc{SynSym} currently excludes metaphorical or figurative expressions, which may constrain the linguistic diversity. Future work will explore generation strategies that can incorporate such diverse expressions while maintaining label fidelity. Furthermore, we aim to explore more advanced methods for symptom identification to fully leverage the potential of synthetic data in diverse clinical contexts.

    \begin{table}[t]
        \caption{Comparison with back-translation augmentation across three benchmark datasets. All models use MentalBERT as the backbone encoder.}
        \small
        \renewcommand{\arraystretch}{1.3}
        \centering
        \resizebox{1.0\linewidth}{!}{
        \begin{tabular}{l|ccc}
        \hline
        \multicolumn{1}{c|}{\multirow{2}{*}{\textbf{Training}}} & 
        \textbf{PsySym} & \textbf{PRIMATE} & \textbf{D2S} \\
        \cline{2-4}
        & Macro-F1 ↑ & Macro-F1 ↑ & Macro-F1 ↑ \\ \hline
        \textbf{Real}\textsubscript{only} & 0.811 ± 0.004 & 0.643 ± 0.031 & 0.603 ± 0.030 \\
        \textbf{Back-trans}\textsubscript{with Real} & 0.817 ± 0.005 & 0.641 ± 0.019 & 0.600 ± 0.034 \\ \hline
        \textbf{\textsc{SynSym}}\textsubscript{with Real} & \textbf{0.830 ± 0.006} & \textbf{0.650 ± 0.016} & \textbf{0.614 ± 0.023} \\ \hline
        \end{tabular}}
        \label{table:backtranslation}
    \end{table}

\section*{Ethical Consideration}
    We carefully considered the ethical implications of generating synthetic data related to psychiatric symptoms. Given the goal of supporting psychiatric symptom identification, some outputs include high-risk content such as suicidal ideation or self-harm to reflect the severity of real-world symptoms. To mitigate potential harm, prompts were carefully designed to generate first-person statements that are clinically meaningful and explicitly describe the speaker’s own symptoms, while avoiding graphic, exaggerated, or emotionally manipulative language, such as content that encourages or romanticizes suicide or self-harm. 

\section*{Disclosure of AI Use}
    We used OpenAI’s ChatGPT to assist with grammar correction and clarity improvement. All technical content, analysis, and writing decisions were made solely by the authors.

\begin{acks}
This research was supported by the MSIT (Ministry of Science and ICT), Korea, under the Global Research Support Program in the Digital Field program (RS-2024-00425354) supervised by the IITP (Institute for Information \& Communications Technology Planning \& Evaluation), and by the MOTIE (Ministry of Trade, Industry and Energy), Korea, under the Competitiveness Reinforcement Project (VCSK2502) for Industrial Clusters supervised by the KICOX (Korea Industrial Complex Corporation).
\end{acks}

\clearpage
\bibliographystyle{ACM-Reference-Format}
\balance
\bibliography{custom}

\appendix
\clearpage
\section{Prompt Templates}
\label{appendix:prompts_framework}
\begin{figure}[h]
        \centering
        \includegraphics[width=1.0\linewidth]{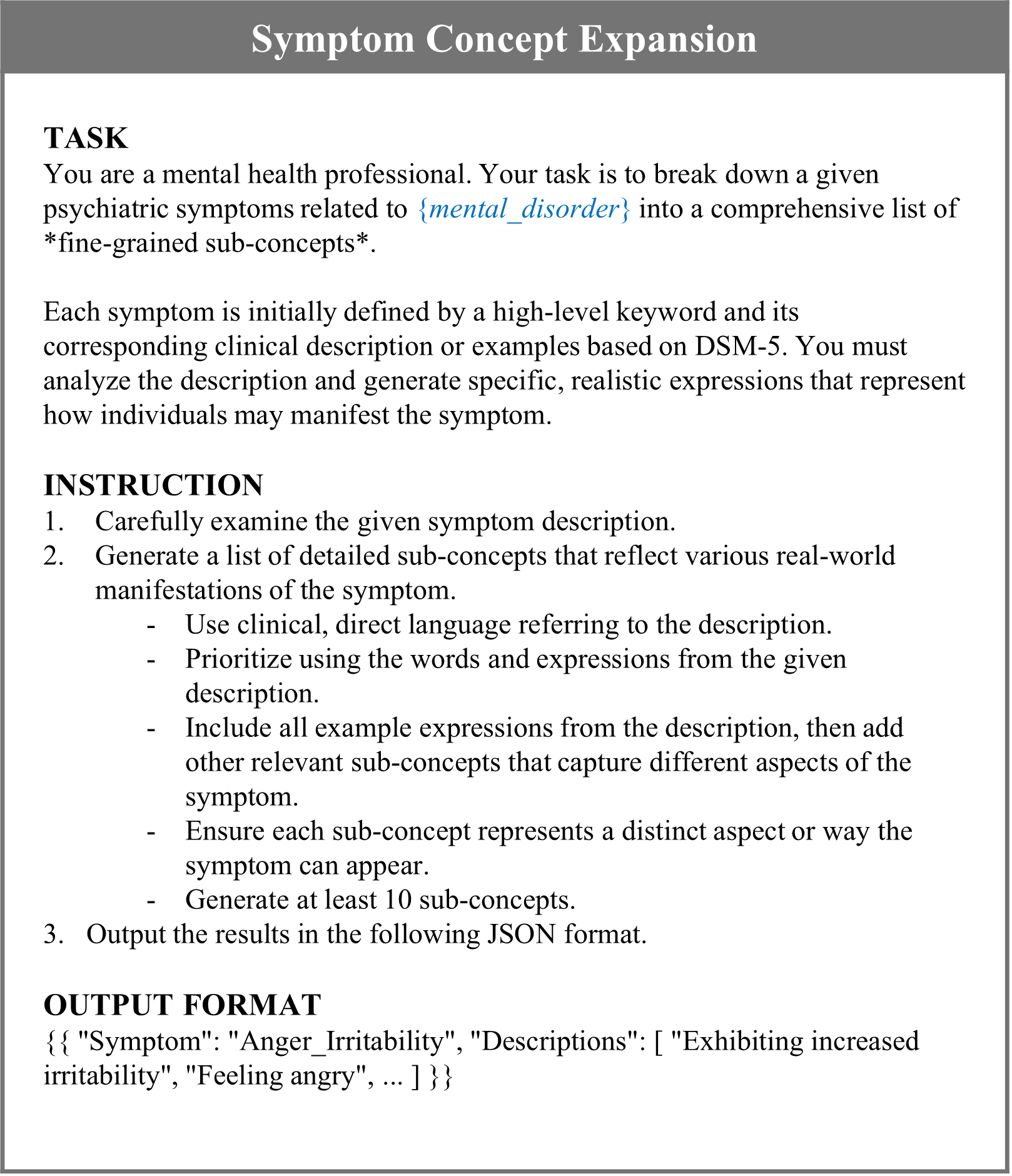}
        \caption{Prompt for symptom concept expansion. Given a symptom keyword and its brief description as input, the LLM is instructed to generate a list of expanded sub-concepts that reflect more fine-grained aspects of the target symptom.}
        \label{fig:prompt_1}
    \end{figure}

\begin{figure}[h]
        \centering
        \includegraphics[width=1.0\linewidth]{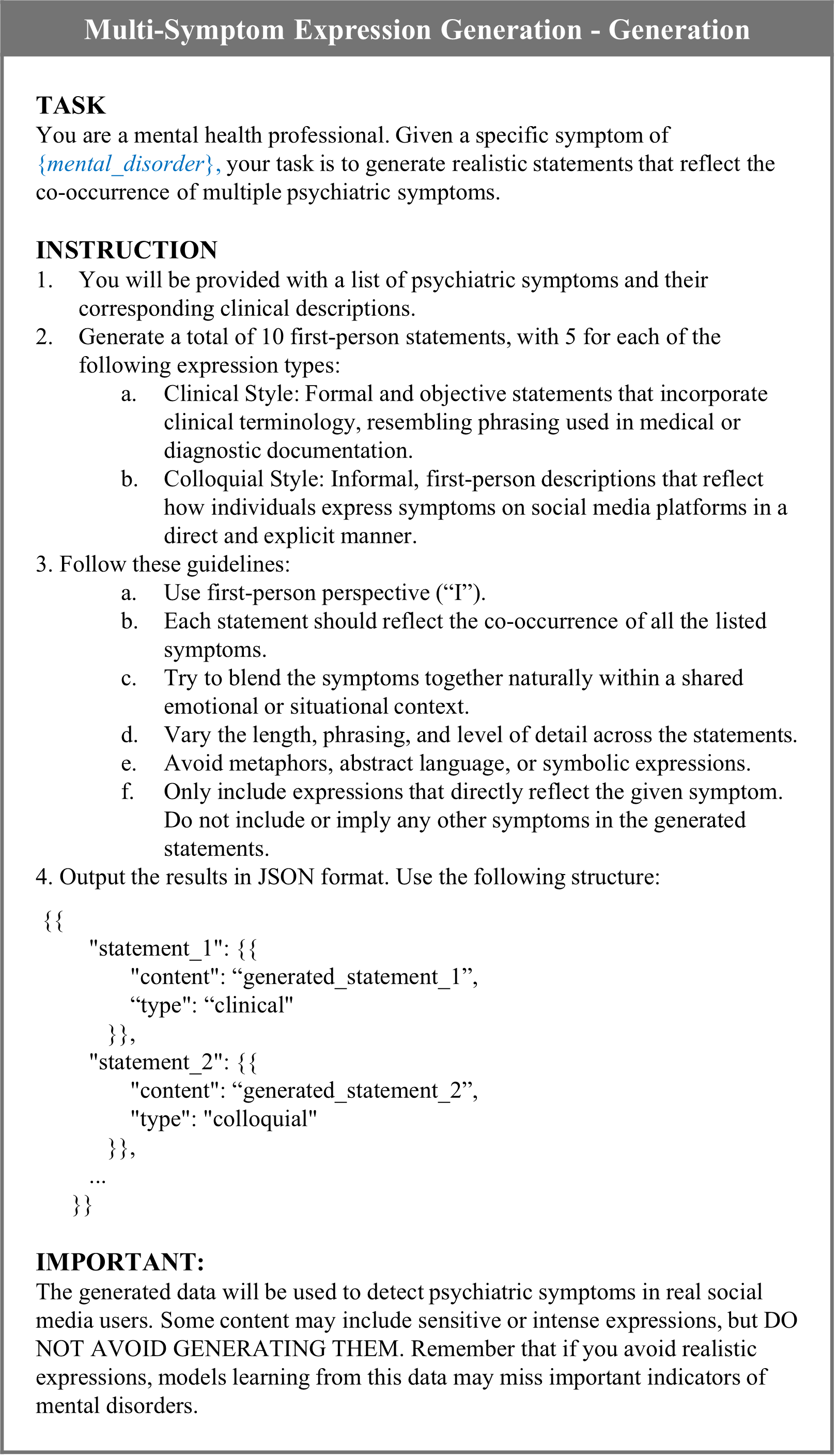}
        \caption{Prompt for multi-symptom expression generation. Given a set of symptom combinations as input, the LLM generates natural language expressions that simultaneously reflect all symptoms.}
        \label{fig:prompt_4}
        \vspace{-0.15in}

\end{figure}

\begin{figure}[]
        \centering
        \includegraphics[width=1.0\linewidth]{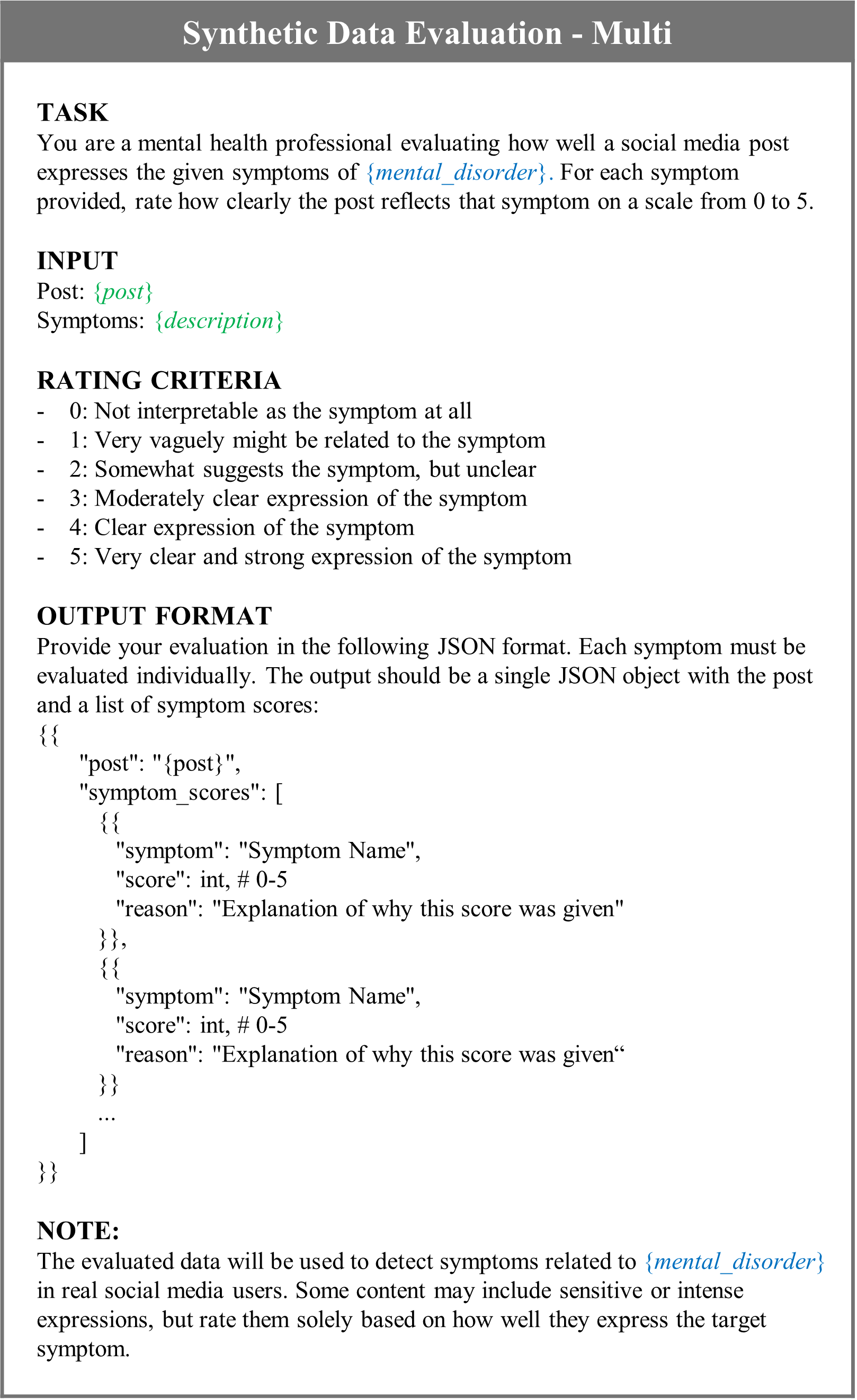}
        \caption{Prompt for evaluating multi-symptom expressions. Given a generated expression and its corresponding symptom labels, the LLM is instructed to assess whether all target symptoms are accurately reflected in the text.}
    \label{fig:prompt_5_2}
    \vspace{-0.1in}
    \end{figure}

We provide the prompt templates used in this work. Figures~\ref{fig:prompt_1}--\ref{fig:prompt_5_2} present the prompt templates employed in the \textsc{SynSym} framework. For symptom concept expansion, we set the temperature parameter to 0.0 to ensure controlled and deterministic generation. During symptom expression generation, we increased the temperature to 0.8 to encourage diversity and creativity in the generated expressions. For the evaluation stage, the temperature was again set to 0.0 to ensure stable and consistent assessments. Additional prompt variants not shown here are available in the released codebase. 

\section{Details of Synthetic Dataset}
\label{appendix:synthetic_dataset}

\subsection{Examples of Symptom Expressions}
To facilitate understanding of the synthetic data generated by \textsc{SynSym}, Table~\ref{tab:examples_synth} presents representative examples for selected symptoms in both clinical and colloquial styles. The clinical style leverages terminology typically used in medical contexts, whereas the colloquial style adopts language more characteristic of social media discourse, maintaining a direct and explicit mode of expression. 

\subsection{Examples of Sub-Concepts}
Table~\ref{table:subconcepts} presents representative examples of sub-concepts generated for each symptom keyword, illustrating how high-level symptoms are decomposed into fine-grained clinical concepts. By leveraging these sub-concepts, \textsc{SynSym} enables the generation of diverse symptom expressions, capturing the linguistic variability inherent in real-world contexts, where the same clinical symptom may be expressed in various ways by different patients.

\subsection{Distribution of Symptom Labels}
Table~\ref{table:label_synthetic} presents the symptom classes included in the synthetic dataset along with the number of examples generated for each class. To reflect the frequency of symptom co-occurrence patterns observed in clinical practice, we generated multi-symptom combinations based on prior clinical literature on depressive symptom networks~\cite{fried2016good, gijzen2021suicide, mullarkey2019using, quinn2023relations}. \textit{Depressed Mood} appears most frequently, as it serves as a core symptom in major depressive disorder and commonly co-occurs with other symptoms.

\section{Details of Experiments}
\subsection{Remapping Labels}
\label{appendix:remapping_labels}

To facilitate cross-dataset evaluation, we remapped DSM-5--based symptom labels to PHQ-9 categories. Most of the 14 DSM-5--derived labels were aligned to PHQ-9 categories in a one-to-one manner based on semantic similarity. For example, \textit{Decreased Energy, Tiredness, Fatigue} was mapped to \textit{Lack of Energy}, and \textit{Depressed Mood} to \textit{Feeling Down}. However, three exceptions were handled with specific rationale. First, \textit{Anger or Irritability} was excluded due to the lack of a semantically equivalent category in PHQ-9. Second, \textit{Genitourinary Symptoms}, which often reflect libido loss, were mapped to \textit{Lack of Interest}. Third, \textit{Inattention}, \textit{Indecisiveness}, and \textit{Poor Memory}, which share cognitive impairment features, were jointly mapped to \textit{Concentration Problems}. The complete mapping is shown in Table~\ref{tab:remapping}.

\begin{table}[h]
    \caption{Label remapping scheme from DSM-5 symptom categories to PHQ-9 symptom categories, used for training and evaluation across datasets with different annotation schemes.}    \centering
    \small
    \label{tab:remapping}
    \begin{tabular}{ll}
    \toprule
    \textbf{DSM-5 Label} & \textbf{PHQ-9 Category} \\
    \midrule
    Decreased Energy or Fatigue & Lack of Energy \\
    Depressed Mood & Feeling Down \\
    Genitourinary Symptoms & Lack of Interest \\
    Hyperactivity or Agitation & Hyper/Lower Activity \\
    Inattention & Concentration Problems \\
    Indecisiveness & Concentration Problems \\
    Poor Memory & Concentration Problems \\
    Suicidal Ideas & Suicidal Ideation \\
    Worthlessness and Guilt & Low Self-Esteem \\
    Loss of Interest or Motivation & Lack of Interest \\
    Pessimism & Feeling Down \\
    Sleep Disturbance & Sleep Disturbance \\
    Weight or Appetite Change & Appetite Change \\
    \textit{Anger or Irritability} & \textit{(excluded)} \\
    \bottomrule
    
    \end{tabular}
\end{table}
    
    \begin{table}[]
    \caption{Label distribution of the synthetic dataset. \textit{Depressed Mood} appears most commonly as a co-occurring symptom.}
    \centering
        \small
        \begin{tabular}{lr}
        \hline
        \textbf{Symptom}                              & \multicolumn{1}{l}{\# \textbf{Samples}} \\ \hline
        Anger Irritability                   & 1294                           \\
        Decreased Energy, Tiredness, Fatigue & 2106                           \\
        Depressed Mood                       & 4829                           \\
        Genitourinary Symptoms               & 1186                           \\
        Hyperactivity Agitation              & 1569                           \\
        Inattention                          & 2054                           \\
        Indecisiveness                       & 1320                           \\
        Suicidal Ideas                       & 2012                           \\
        Worthlessness and Guilt              & 2294                           \\
        Loss of Interest or Motivation       & 2703                           \\
        Pessimism                            & 1958                           \\
        Poor Memory                          & 1374                           \\
        Sleep Disturbance                    & 2981                           \\
        Weight and Appetite Change           & 1993                           \\ \hline
        \end{tabular}
    \label{table:label_synthetic}
    \end{table}

\subsection{Hyperparameters}
\label{appendix:hyperparameters}
The experiments were conducted on two NVIDIA Quadro RTX A5000 GPUs. For training on the synthetic data, we employed the AdamW optimizer with a default weight decay of 0.01 and a learning rate of 5e-5. During additional training on real benchmark datasets, the learning rate was reduced to 3e-5 to facilitate more stable adaptation. The maximum sequence length was set to 512 tokens, and the batch size was set to either 32 or 64, depending on the input sequence length.

\section{Symptom-level Analysis Results}
\label{appendix:experimental_result}
\begin{figure*}[]
    \centering
    \includegraphics[width=\textwidth]{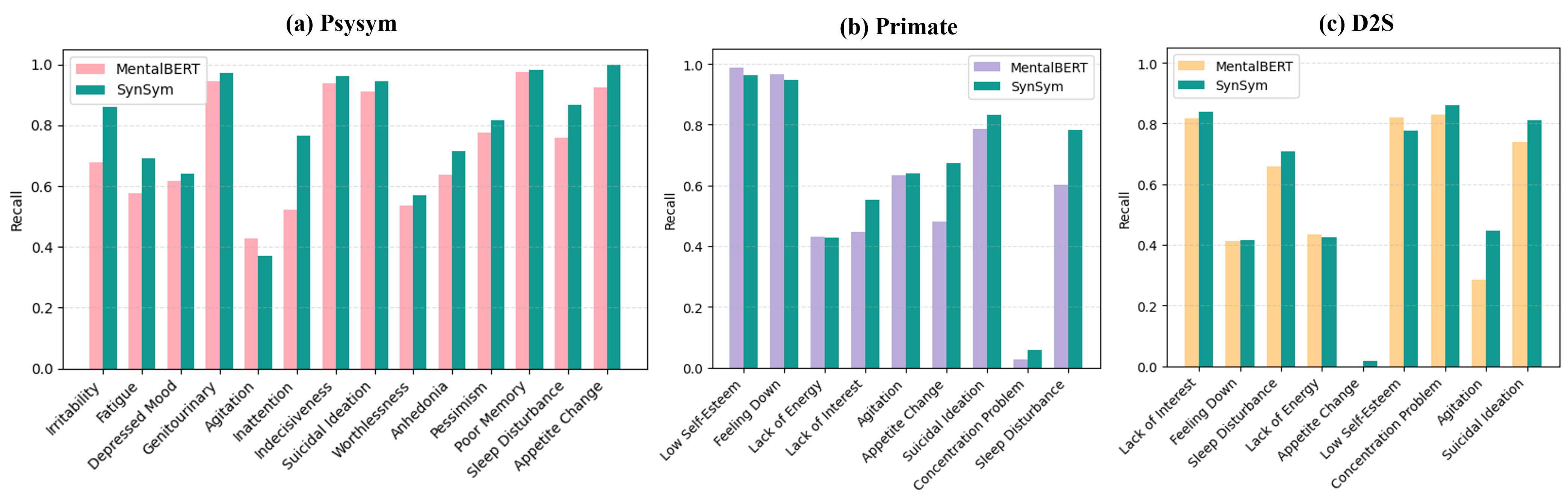}
    \caption{Symptom-level comparison between \textsc{SynSym}\textsubscript{with Real} and MentalBERT on three benchmark datasets: PsySym (left), PRIMATE (center), and D2S (right). \textsc{SynSym} shows higher recall across most symptoms, particularly for low-frequency or underperforming symptom classes.}
    \label{fig:visualization}
\end{figure*}

Figure~\ref{fig:visualization} shows symptom-level performance analysis between MentalBERT trained solely on real data and \textsc{SynSym}\textsubscript{with Real} across three datasets. The results demonstrate consistent improvements across most symptoms, particularly on PsySym, which is notable for previously underrepresented symptoms with low baseline performance, such as \textit{Inattention}. However, performance on certain symptoms remained low, particularly \textit{Appetite Change} in D2S and \textit{Concentration Problems} in PRIMATE. We attribute this to dataset-specific differences in annotation schemes and expression styles, suggesting the need for standardized symptom annotation protocols and methodological extensions to encompass more diverse linguistic styles.

\begin{table*}[]
    \caption{Representative examples of synthetic symptom expressions generated by \textsc{SynSym}, illustrating both clinical-style and colloquial-style expressions for selected symptoms.}
    \renewcommand{\arraystretch}{1.1}
    \centering
    \resizebox{1.0\linewidth}{!}{
    \begin{tabular}{l|l|l}
    \hline
    \textbf{Symptom} & \textbf{Type} & \textbf{Example} \\ \hline
    \multirow{2}{*}{\textbf{Depressed Mood}} & Clinical & There is a persistent and pervasive sadness in my emotional state. \\
     & Colloquial & I'm not feeling anything at all. \\ \hline
    \multirow{2}{*}{\textbf{Suicidal Ideation}} & Clinical & I frequently have intrusive thoughts about ending my life. \\
     & Colloquial & I keep thinking that the only way to escape this feeling of being stuck is to end it all. \\ \hline
    \multirow{2}{*}{\textbf{Genitourinary Symptoms}} & Clinical & \multicolumn{1}{r}{I frequently exhibit nocturia, waking up multiple times throughout the night specifically to urinate.} \\
     & Colloquial & Sometimes I just pee myself without being able to stop it. \\ \hline
    \end{tabular}}
    \label{tab:examples_synth}

\end{table*}
\begin{table*}[]
\caption{Representative examples of sub-concepts generated for selected symptoms, illustrating plausible manifestations of high-level symptom concepts in real-world contexts.}
\centering
\small
\renewcommand{\arraystretch}{1.05}
\begin{tabular}{p{5cm} p{12cm}}
\hline
\textbf{Symptom} & \textbf{Representative Sub-Concept Examples} \\ \hline
\textbf{Depressed Mood} &
- Persistent feelings of sadness \newline
- Experiencing frequent crying spells \newline
- Feeling empty \\ \hline
\textbf{Suicidal Ideas} &
- Making a suicide attempt \newline
- Experiencing fear of dying \newline
- Having recurrent thoughts of death \\ \hline
\textbf{Inattention} &
- Diminished ability to concentrate \newline
- Avoids tasks that require sustained mental effort \newline
- Fails to finish duties in the workplace \\ \hline
\textbf{Sleep Disturbance} &
- Difficulty falling asleep \newline
- Restless, unsatisfying sleep \newline
- Insomnia \\ \hline
\textbf{Loss of Interest or Motivation} &
- Lose motivation to do things \newline
- Feelings of detachment or estrangement from others \newline
- Reduced desire to pursue personal goals or ambitions \\ \hline
\textbf{Worthlessness and Guilt} &
- Feeling disappointed in oneself \newline
- Expecting to be punished \newline
- Experiencing guilty preoccupations or ruminations over minor past failings \\ \hline
\textbf{Decreased Energy, Tiredness, Fatigue} &
- Being easily fatigued \newline
- Feeling a persistent lack of energy \newline
- Experiencing a constant sense of tiredness \\ \hline
\textbf{Weight and Appetite Change} &
- Eating much more rapidly than normal \newline
- Feeling unable to stop eating or control what or how much is being eaten \newline
- Experiencing physical discomfort or pain due to overeating \\ \hline
\end{tabular}
\label{table:subconcepts}
\end{table*}

\end{document}